**ISSN 2220-9182**# DME-RR Department of Mechanical Engineering Research Report

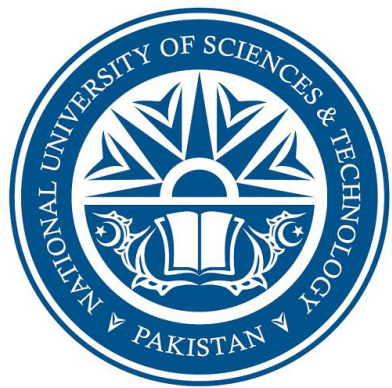

*Department of Mechanical Engineering,*
*College of Electrical and Mechanical Engineering,*
*National University of Sciences and Technology,*
*Islamabad-44000, Pakistan*

**2011-02**

**Department of Mechanical Engineering-Research Report (DME-RR)**

# Designing a Miniature Wheel Arrangement for Mobile Robot Platforms

*by*

**Saheeb Ahmed Kayani**

## Table of Contents

# List of Illustrations

# Abstract

In this research report details of design of a miniature wheel arrangement are presented. This miniature wheel arrangement is essentially a direction control mechanism intended for use on a mobile robot platform or base. The design is a specific one employing a stepper motor as actuator and as described can only be used on a certain type of wheeled robots. However, as a basic steering control element, more than one of these miniature wheel arrangements can be grouped together to implement more elaborate and intelligent direction control schemes on varying configurations of wheeled mobile robot platforms.

# Background Information

Indigenously developed mobile robot platforms are commonly used by students (and professionals) in Pakistan for different robotic competitions. These platforms come in different configurations and are formed out of sheet metal (usually Aluminum). The design may vary from one platform to the other but the basic purpose remains the same i.e. mobility. Although complicated mechanisms employing gear-trains and chains are also used, but most often in wheeled robots the motors are directly coupled to the wheels and any difference in the speeds of the motors is compensated through control electronics and programming. In this type of drive systems turning and turning-control offers a challenge. When a turn is executed usually one motor is stopped suddenly (and the wheel attached to it acts as a pivot) while the other runs forwards or backwards to achieve the desired turning effect. In this manner smooth turning may not be possible (especially if the robot is following/tracking a line). Also in some cases the robot has to be stopped completely to make a turn or one of the motors is spinned the other way to suddenly change the direction of the moving robot. (Spinning is required to save the machine from overshooting the turning point). This may induce slippage and also unnecessary control (programming) complexity.

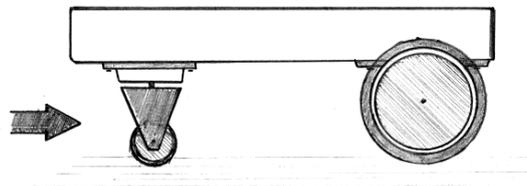

Fig. 1 Schematic of the mobile robot platform. (Not to scale.)

Similar problems were encountered while designing a robot platform for a fire fighting robot design competition in 2003.

As can be seen from Fig. 1, the robot platform used a direct drive system but for ease of turning a specially designed mechanism was employed in the front end of the platform coupled with a stepper motor for incremental direction control. By simultaneously controlling the stepper motor shaft angle (of turn) and the speed of the drive motors, the robot was able to navigate around all bends of the track satisfactorily.

## Description of Design and Operation

In this section, certain details about different aspects of mechanical design and operation of the miniature wheel arrangement have been included. The miniature wheel arrangement can be divided into three distinct parts:

- (a) Main frame made of mild steel, Fig. 2a(k).
- (b) Wheel made of aluminum alloy with tire, Fig. 3a.
- (c) Jaw for holding stepper motor shaft made of oil hard steel, Fig. 2a(h).

The main frame of the miniature wheel arrangement has been developed from a mild steel sheet of 1 mm thickness. The sheet is bent at right angles at the top, leaving a gap of 10 mm (Fig. 2b(c)) between inner faces of the two bent sides. These opposing sides are cut at angles of 30° (Fig. 2a(g)) inwards to produce a downward tapered shape as seen in Fig. 2a. Two identical holes on both sides of the main frame (Fig. 2a(f)) with diameter 3 mm are drilled to accommodate the aluminum alloy wheel shaft.

The wheel and tire combination (Figs. 3a and 3b) is the most important of all parts in the miniature wheel arrangement. The wheel has been machined from a single section of aluminum alloy billet. A continuous circumferential recess or notch (Figs. 4a and 4b(a)) 2 mm deep and 1 mm in length at base (Fig. 4a(a)) is machined to provide a firm holding for the polymer tire. This notch is slanted at an angle of 53° (Fig. 4b(d)) on both ends. This angle is calculated using the dimensions of sides of the right angle triangle formed after the notch was created (Fig. 4a). Using data from Fig. 4b we can calculate:

$a^2 + b^2 = c^2$
$(2)^2 + (1.5)^2 = c^2$
$c^2 = 6.25$
$c = 2.5$ mm
$b = c\cos\theta$
$1.5 = 2.5\cos x$

*x* = 53.1°

(*x* + *y* + 90° = 180°)

*x* = 53°

*y* = 127° (Fig. 4b(e))

Collars on both sides (Fig. 3a(b)) of diameter 6 mm and breadth 1 mm provide not only structural support to the wheel but also ensure smooth motion when it is assembled onto the main frame (Figs. 2a and 2b). When assembled with the main frame the wheel rests on an aluminum alloy shaft (Fig. 2a(f)) of diameter 3 mm whose ends are slightly tapped to serve as pins, holding it in place. The wheel has been machined with strict dimension control and when assembled a clearance (Fig. 2b(k)) of 1 mm is maintained on both sides to allow free rotation under various loading and movement conditions.

On top of the main frame a hollow jaw (Fig. 2a(h)) is provided, machined out of oil hard steel for holding the stepper motor shaft in place with the help of a small screw (Fig. 2a(a)) of diameter 3.5 mm. This screw is operated with a hex key wrench (1.5 mm) and slides in and out of the threaded hole especially machined on the front side of the jaw. This jaw is joined with the main frame through gas welding. The thickness of the welding metal layer is accounted for while measuring the complete length (42 mm) of the miniature wheel arrangement. As the diameter of this jaw is slightly greater than the breadth of the main frame, a lateral offset (Figs. 2b(e) and 2c(a)) is created on both sides of the upper end of the main frame.

The operation of the miniature wheel arrangement is simplified due to the undemanding mechanical design described above. First the stepper motor is turned on, its shaft turns and attains its predefined initial position. The shaft is then slid into the jaw and the screw (Fig. 2a(a)) is turned by the hex key wrench until it firmly grips the stepper motor shaft inside the jaw. While doing this the front side (Fig. 2b) of the miniature wheel arrangement is always kept inline with the forward direction that is to be followed by the mobile robot (platform). The robot is then moved a few paces on a flat surface to ensure that it rolls in a straight line. If not so, any slight adjustment required in the orientation of the miniature wheel arrangement is performed by opening the screw and repeating the above process until the robot is able to achieve a straight-line motion.

A little lubrication may be needed on the wheel shaft to avoid any undue friction and subsequent damage. Also it is assumed that maximum weight of the robot is supported by the drive wheels located at the rear end and only a minor component of it is transferred to the miniature wheel arrangement. A unique feature of this design is that more than one of these miniature wheel arrangements can be grouped together for more controlled and systematic steering.

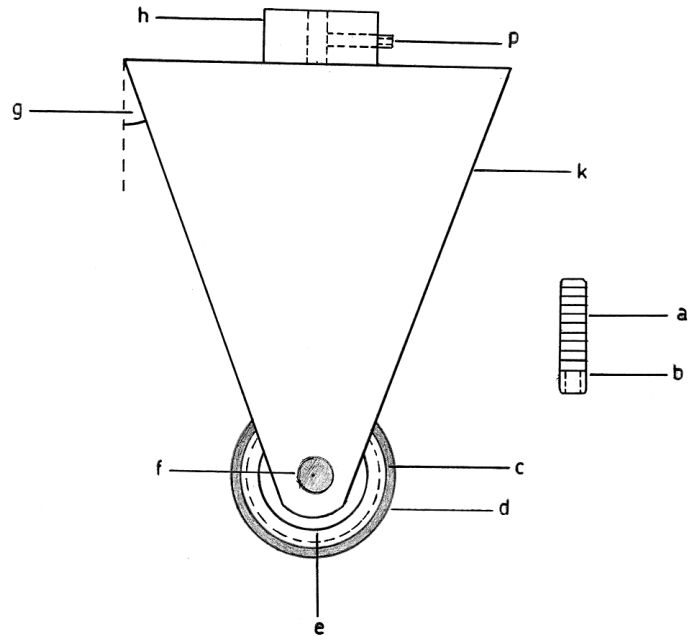

Fig. 2a Miniature wheel arrangement (side view).

| Parameter | Remarks | Dimension |
|---|---|---|
| a | Pin used to grip stepper motor shaft | Diameter 3.5 mm; threaded with 10 threads on 7 mm length |
| b | End hollowed for accommodating hex key wrench | |
| c | Wheel | Diameter 16 mm |
| d | Tire | Diameter 20 mm |
| e | Collar | Diameter 6 mm |
| f | Wheel shaft | Diameter 3 mm |
| g | | $z = 30°$ |
| h | Hollow jaw for providing gripping support to stepper motor shaft | |
| k | Main frame of the miniature wheel arrangement | Length 29 mm |
| p | See a and b above | |

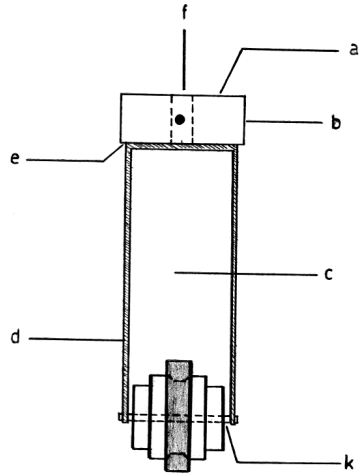

Fig. 2b Miniature wheel arrangement (front view).

| Parameter | Remarks | Dimension |
|---|---|---|
| a | See Fig. 2a(h) | Diameter 14 mm |
| b | See Fig. 2a(h) | Length 7.5 mm |
| c | Breadth of the main frame | 12 mm |
| d | Thickness of plate | 1 mm |
| e | Lateral offset | 1 mm |
| f | Hole for accommodating stepper motor shaft | Diameter 7 mm |
| k | Clearance for free movement of wheel | 1 mm |

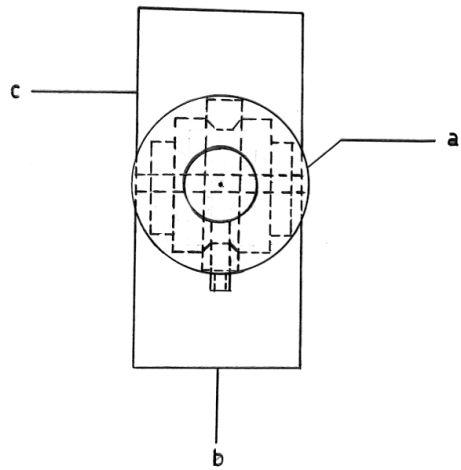

Fig. 2c Miniature wheel arrangement (top view).

| Parameter | Remarks | Dimension |
|---|---|---|
| a | Lateral offset | 1 mm |
| b | See Fig. 2a(k) | Breadth 12 mm |
| c | See Fig. 2a(k) | Width 30 mm |

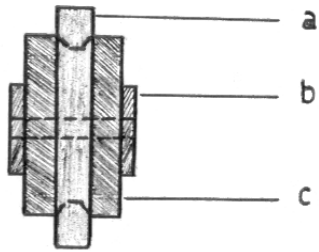

Fig. 3a Wheel and tire combination (front view).

| Parameter | Remarks | Dimension |
|---|---|---|
| a | Tire | Diameter 20 mm; cross-section thickness 4 mm |
| b | Collar | Diameter 6 mm; Breadth 1 mm |
| c | Wheel | Diameter 16 mm; Breadth 1 mm |

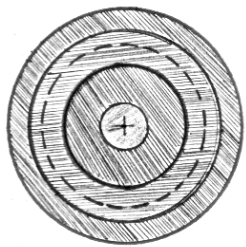

Fig. 3b Wheel and tire combination (side view).

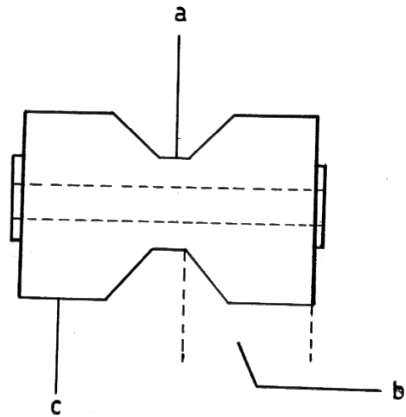

Fig. 4a Wheel notch geometry: design (top view).

| Parameter | Remarks | Dimension |
|---|---|---|
| a | Notch | Length (at base) 1 mm |
| b |  | 1.5 mm |
| c |  | 1 mm |

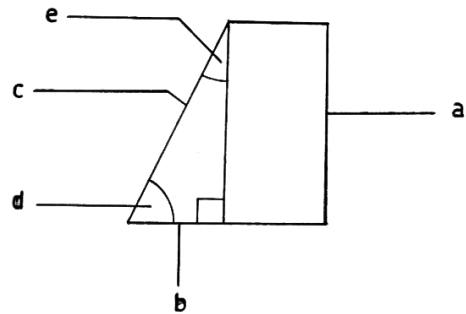

Fig. 4b Wheel notch geometry: calculation.

| Parameter | Dimension |
|---|---|
| a | 2 mm |
| b | 1.5 mm |
| c | 2.5 mm |
| d | $x = 53°$ |
| e | $y = 127°$ |

## Acknowledgment

This project was carried out as part of a fire fighting robot design competition (now National Engineering Robotics Contest) in fall of 2003. The project team included GC Tasleem, GC Akhtar, GC Jahanzeb, and NS Saheeb of $22^{nd}$ degree engineering course in department of mechatronics engineering of NUST College of E&ME, Rawalpindi. Although the robot (named Tornado) design was a team effort, the miniature wheel arrangement described in this research report was developed solely by Mr. Saheeb Ahmed Kayani.

The fabrication of the miniature wheel arrangement was carried out at New Light Engineering Works, Railway Road, Rawalpindi. All materials used in the fabrication process were either provided by the machine shop already mentioned or procured at Tall Man Industrial Products, Kashmir Road, Rawalpindi.